\documentclass[runningheads]{llncs}

\usepackage{eccv}

\usepackage{eccvabbrv}

\usepackage{graphicx}
\usepackage{booktabs}

\usepackage[accsupp]{axessibility}  

\usepackage{hyperref}
\hypersetup{
    colorlinks=true,
    linkcolor=black,
    filecolor=black,
    citecolor=black,
}
\usepackage{orcidlink}
\usepackage{graphicx}
\usepackage{amsmath}
\usepackage{amssymb}
\usepackage{booktabs}
\usepackage{multirow}
\usepackage{bbding}
\usepackage{xcolor}
\begin{document}

\title{Domain-adaptive Video Deblurring via Test-time Blurring} 

\titlerunning{Domain-adaptive Video Deblurring via Test-time Blurring}

\author{Jin-Ting He\inst{*, 1}\and
Fu-Jen Tsai\inst{*, 2}\and
Jia-Hao Wu\inst{1}\and
Yan-Tsung Peng\inst{3}\and
Chung-Chi Tsai\inst{4}\and
Chia-Wen Lin\inst{2}\and
Yen-Yu Lin\inst{1}}

\authorrunning{J.-T. He et al.}

\institute{National Yang Ming Chiao Tung University, Taiwan\\
\email{jinting.cs12@nycu.edu.tw, jiahao.11@nycu.edu.tw, lin@cs.nycu.edu.tw}\and
National Tsing Hua University, Taiwan\\
\email{fjtsai@gapp.nthu.edu.tw, cwlin@ee.nthu.edu.tw}\and
National Chengchi University, Taiwan\\
\email{ytpeng@cs.nccu.edu.tw}\and
Qualcomm Technologies, Inc., San Diego\\
\email{chuntsai@qti.qualcomm.com}
}

\maketitle

\begingroup
\renewcommand\thefootnote{}\footnotetext{* equal contribution}
\addtocounter{footnote}{-1}
\endgroup

\begin{abstract}
Dynamic scene video deblurring aims to remove undesirable blurry artifacts captured during the exposure process. Although previous video deblurring methods have achieved impressive results, they suffer from significant performance drops due to the domain gap between training and testing videos, especially for those captured in real-world scenarios.
%
To address this issue, we propose a domain adaptation scheme based on a blurring model to achieve test-time fine-tuning for deblurring models in unseen domains. Since blurred and sharp pairs are unavailable for fine-tuning during inference, our scheme can generate domain-adaptive training pairs to calibrate a deblurring model for the target domain.
%
First, a Relative Sharpness Detection Module is proposed to identify relatively sharp regions from the blurry input images and regard them as pseudo-sharp images. 
Next, we utilize a blurring model to produce blurred images based on the pseudo-sharp images extracted during testing. To synthesize blurred images in compliance with the target data distribution, we propose a Domain-adaptive Blur Condition Generation Module to create domain-specific blur conditions for the blurring model. Finally, the generated pseudo-sharp and blurred pairs are used to fine-tune a deblurring model for better performance. 
Extensive experimental results demonstrate that our approach can significantly improve state-of-the-art video deblurring methods, providing performance gains of up to 7.54dB on various real-world video deblurring datasets. The source code is available at \href{https://github.com/Jin-Ting-He/DADeblur}{https://github.com/Jin-Ting-He/DADeblur}.

\end{abstract}
\section{Introduction}
Videos captured in dynamic scenes often appear blurred and have blurry artifacts due to camera shaking or moving objects captured during the exposure process. The unwanted blur severely degrades visual qualities and reduces the accuracy of subsequent computer-vision applications. Dynamic scene video deblurring aims to restore such blurred videos, considered a highly ill-posed problem due to the unknown varying blurs.  

Video deblurring has made remarkable progress with the development of deep learning. Numerous methods~\cite{Pan_2020_CVPR,Li_2021_CVPR,zhou2021rta,zhong2022real,zhang2022spatio,wang2022MMP,Suin_2021_CVPR,chao2022,Ji_2022_CVPR,Jiang_2022_ECCV,nah2019recurrent,su2017deep,liang2022vrt} have applied Convolutional Neural Networks (CNNs) for video deblurring. Among these methods, Recurrent Neural Network-based (RNN-based) methods~\cite{Ji_2022_CVPR,chao2022,wang2022MMP,zhong2022real} are commonly employed to extract spatio-temporal features. To better capture motion cues in videos, several methods~\cite{Pan_2020_CVPR,Li_2021_CVPR,su2017deep} compute optical flows to get additional motion information for alignment. In addition, motivated by the success of Transformer in computer vision~\cite{dosovitskiy2020vit,DETR,Ranftl21,chu2021Twins,chen2022regionvit,zhu2021deformable,xie2021segformer}, Transformer-based methods~\cite{liang2022rvrt,liang2022vrt,fgst} have been proposed to model long-range, spatio-temporal dependency. 
Since these methods have been dedicated to improving deblurring through architectural designs, their performance often drops substantially when a domain gap exists between training and testing videos. 

\begin{figure}[t!]
\begin{center}
\includegraphics[width=1\columnwidth]{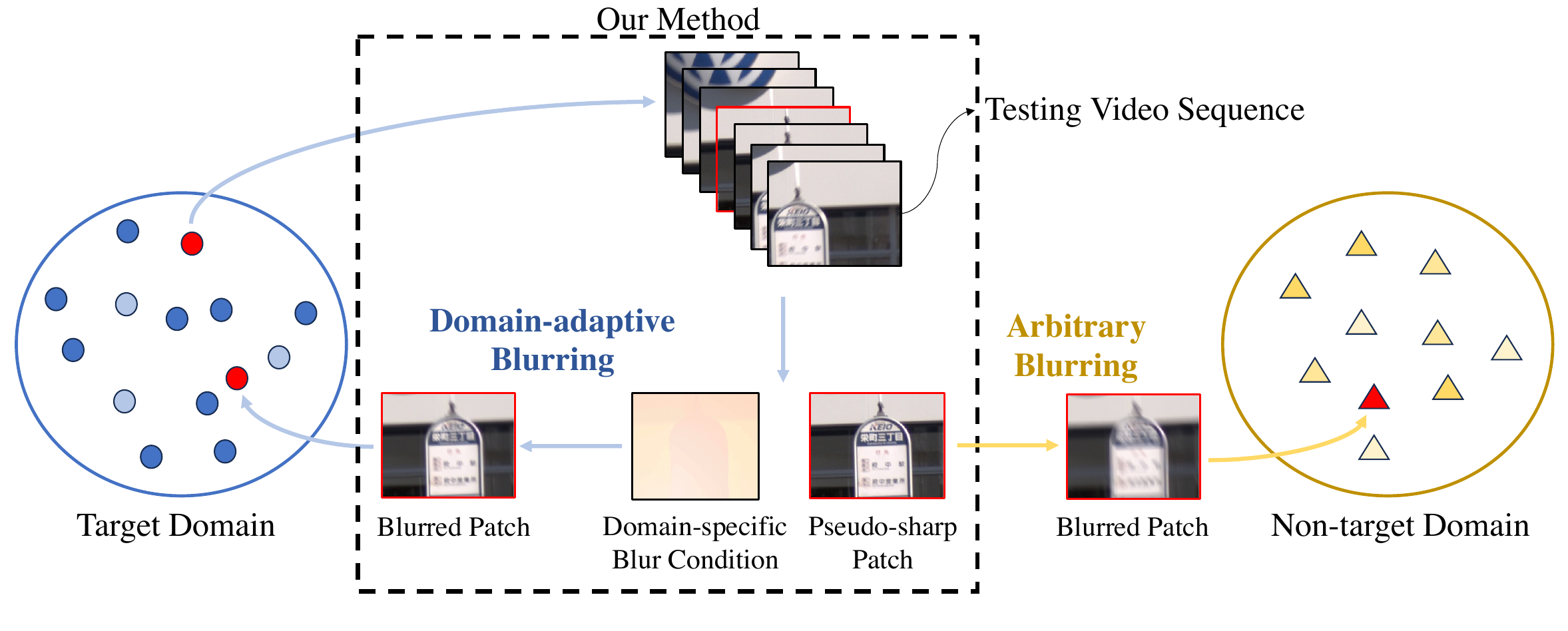}
\end{center}
\caption{Illustration of the proposed domain adaptation method. It can generate domain-specific blur conditions for a blurring model to produce blurred video frames from the chosen pseudo-sharp patch. These blurred frames can be used to fine-tune video deblurring models and improve their performance in the target domain. 
}
\label{fig:teaser}
\end{figure}

In real-world cases, cameras may capture blurry videos due to different settings and scenarios, such as shutter speed, aperture, or light sources, leading to different blurry patterns with various orientations and magnitudes. Previous deblurring models often face challenges when dealing with blurry patterns not seen during training. However, few methods discuss the domain gap issue in video deblurring. Liu \etal~\cite{Liu_2022_BMVC} attempted to address it via a blurring network that generates blurred images to enable meta-learning~\cite{finn2017model} for adaptation. Even though it could generate blurred data for adaptation, it did not consider helpful temporal information that exists in consecutive frames for domain adaptation.    
%
That is, continuous motion across consecutive frames reveals motion blur trajectories during capturing, and the degree of blurriness implies the blur intensities during exposure. The information implicitly conveys domain-specific cues regarding blur orientations and magnitudes in unseen blurry videos. Therefore, motivated by this observation, we further explore the cues to enhance video deblurring models in unseen domains.    

In this paper, we propose a domain adaptation method for video deblurring, which relies on a blurring model to generate domain-adaptive training pairs for adaptation. The generated data can be used to calibrate deblurring models in unseen domains. The proposed framework is illustrated in Figure~\ref{fig:teaser}. Since blurred and sharp pairs are unavailable for fine-tuning during inference, our scheme can extract relatively sharp regions from blurry videos using the proposed Relative Sharpness Detection Module (RSDM). These relatively sharp regions can be considered pseudo-sharp images. 
Inspired by the recent diffusion-based blurring method, ID-Blau~\cite{ID-Blau}, which can generate blurred images based on a sharp image and arbitrarily specified blur conditions, we adopt ID-Blau as the blurring backbone. Nevertheless, randomly generated blur conditions are not consistent with blur patterns present in test videos. Directly applying ID-Blau with such conditions for blurring and fine-tuning a deblurring model may not help.
Considering blurry videos implicitly provide coherent motion blur cues, we propose a Domain-adaptive Blur Condition Generation Module (DBCGM) containing a Blur Orientation Estimator and Blur Magnitude Estimator to create domain-specific blur conditions tailored for blurring those pseudo-sharp images using ID-Blau. 
It turns out that the generated domain-specific blur pairs can be used to fine-tune deblurring models, thereby achieving domain adaptation during inference. 
%
Our contributions can be summarized as follows:

\begin{itemize}
  \item We propose a test-time domain adaptation method for video deblurring based on ID-Blau, which can generate domain-specific blur conditions to achieve test-time fine-tuning for deblurring models.
  \item We propose a Relative Sharpness Detection Module to detect relatively sharp regions as pseudo-sharp images and a Domain-adaptive Blur Condition Generation Module to generate domain-specific blur conditions for blurring those pseudo-sharp images.
  \item Experimental results demonstrate that our method can significantly enhance state-of-the-art video deblurring models~\cite{zhong2022real,wang2022MMP,Li_2023_CVPR,Pan_2023_CVPR} on five real-world video deblurring datasets, including RealBlur~\cite{rim_2020_ECCV}, RBVD~\cite{chao2022} and three versions of BSD~\cite{zhong2022real}.
\end{itemize}

\section{Related Work}

\subsubsection{Video Deblurring}
Convolutional neural networks have significantly advanced the video deblurring task. Compared to image deblurring~\cite{Nah_2017_CVPR,tao2018srndeblur,Zhang_2019_CVPR,gao2019dynamic,SAPN2020,MT_2020_ECCV,RADN_2020_ECCV,Kupyn_2019_ICCV,Tsai2022Stripformer,Tsai2022BANet}, video deblurring~\cite{Pan_2020_CVPR,Li_2021_CVPR,zhou2021rta,zhong2022real,zhang2022spatio,wang2022MMP,Suin_2021_CVPR,chao2022,Ji_2022_CVPR,Jiang_2022_ECCV,nah2019recurrent,su2017deep,liang2022vrt} can utilize spatial and temporal information from videos to achieve better performance. Recently, several methods adopted RNN-based models~\cite{zhong2022real,wang2022MMP,chao2022,liang2022rvrt,nah2019recurrent,Pan_2023_CVPR,Li_2023_CVPR} to leverage spatio-temporal information in videos. Zhong~\etal~\cite{zhong2022real} proposed an efficient spatio-temporal RNN architecture to catch spatially and temporally varying blurs with RNN cells. Wang~\etal~\cite{wang2022MMP} utilized a motion magnitude prior as guidance to improve deblurring performance. Pan~\etal~\cite{Pan_2023_CVPR} developed wavelet-based feature propagation to transfer features in frequency space recurrently. Li~\etal~\cite{Li_2023_CVPR} proposed a grouped spatial-temporal shift operation to aggregate spatio-temporal features efficiently. 

In addition to progressively propagating features through RNNs, several studies have explored Transformer-based architectures~\cite{liang2022vrt,fgst,liang2022rvrt} to garner long-range information for deblurring. Liang~\etal~\cite{liang2022rvrt} proposed a recurrent video restoration transformer with guided deformable attention. Lin~\etal~\cite{fgst} introduced a flow-guided sparse transformer that leverages optical flows to guide the transformer's attention mechanism. Although these architectural designs for RNNs and Transformers were used to improve video deblurring, they often do not perform up to par when dealing with blurry videos on unseen domains, especially for videos captured in the real world.           

\subsubsection{Domain Adaptation}
Domain adaptation aims to bridge the domain gap between the training set and testing set, which has been widely discussed in vision tasks, such as object detection~\cite{Zhang2023CVPR,li2022cross,li2022sigma} and semantic segmentation~\cite{NEURIPS2022_61aa5576,lai2022decouplenet,hoyer2022hrda}. Nevertheless, little work has been done~\cite{Chi_2021_CVPR,nah2021clean,Liu_2022_BMVC} to address the domain gap issue in deblurring, especially for video deblurring~\cite{Liu_2022_BMVC}. Since we solely have blurred inputs during inference, no ground truth can be used for test-time deblurring calibration. Some methods apply self-supervised strategies to enable fine-tuning during testing. Chi~\etal~\cite{Chi_2021_CVPR} utilized a reconstruction branch to reset its deblurred result to the original blurry input. Nah~\etal~\cite{nah2021clean} employed a fixed reblurring model as a reblurring loss to supervise the deblurring result. Although these methods could update and improve deblurring models through self-supervised learning, they ignore domain-specific blur information in the test domain, which is beneficial to deblurring performance in unseen domains. To generate blurred images with blur characteristics on unseen domains, Liu~\etal~\cite{Liu_2022_BMVC} utilized a GAN to supervise a blurring model to generate blurred images from sharp ones for fine-tuning. In contrast, our method explores domain-specific blur cues from consecutive testing video frames for domain adaptation.

\subsubsection{Diffusion Models}
Diffusion models have demonstrated a strong ability to synthesize realistic images in image generation~\cite{NEURIPS2020_4c5bcfec,song2020denoising,NEURIPS2021_49ad23d1}. Based on the diffusion model, several studies~\cite{nichol2021glide,saharia2022photorealistic,ramesh2021zero,voynov2023sketch,meng2021sdedit,ruiz2023dreambooth,ma2023subject,liang2024control,zhang2023adding,ID-Blau} have utilized it with various conditions to generate controllable results. Some methods~\cite{nichol2021glide,saharia2022photorealistic} used text prompts~\cite{devlin2018bert,2020t5,radford2021learning} in text-to-image generation. Although text embedding provides additional information to generate controllable results, it often has a limitation in spatially guiding the generated images. 
Recently, several studies proposed to spatially guide diffusion models with various prompts~\cite{voynov2023sketch,meng2021sdedit,ruiz2023dreambooth,ma2023subject,liang2024control,zhang2023adding,ID-Blau}. Zhang~\etal\cite{zhang2023adding} proposed Control-Net to fine-tune a pre-trained diffusion model with various prompts by 
zero convolution layers, making the diffusion model spatially controllable without requiring high computational costs. Liang~\etal\cite{liang2024control} proposed a controllable multi-modal image coloring method based on a content-guided deformable autoencoder.
Wu~\etal\cite{ID-Blau} proposed ID-Blau, a diffusion-based blurring model that can generate diverse blurred images based on a sharp image and arbitrarily-specified pixel-wise blur conditions.  
Motivated by the success of ID-Blau, we adopt ID-Blau as a blurring model to generate blurred images during testing. However, simply applying ID-Blau with randomly generated blur conditions would not improve a deblurring model for blurred images coming from different domains. Therefore, we aim to leverage the inherent motion blur cues in testing videos to generate domain-specific blur conditions for adaptation, thereby significantly boosting deblurring performance on unseen domains.

\begin{figure}[t!]
\begin{center}
\includegraphics[width=1\columnwidth]{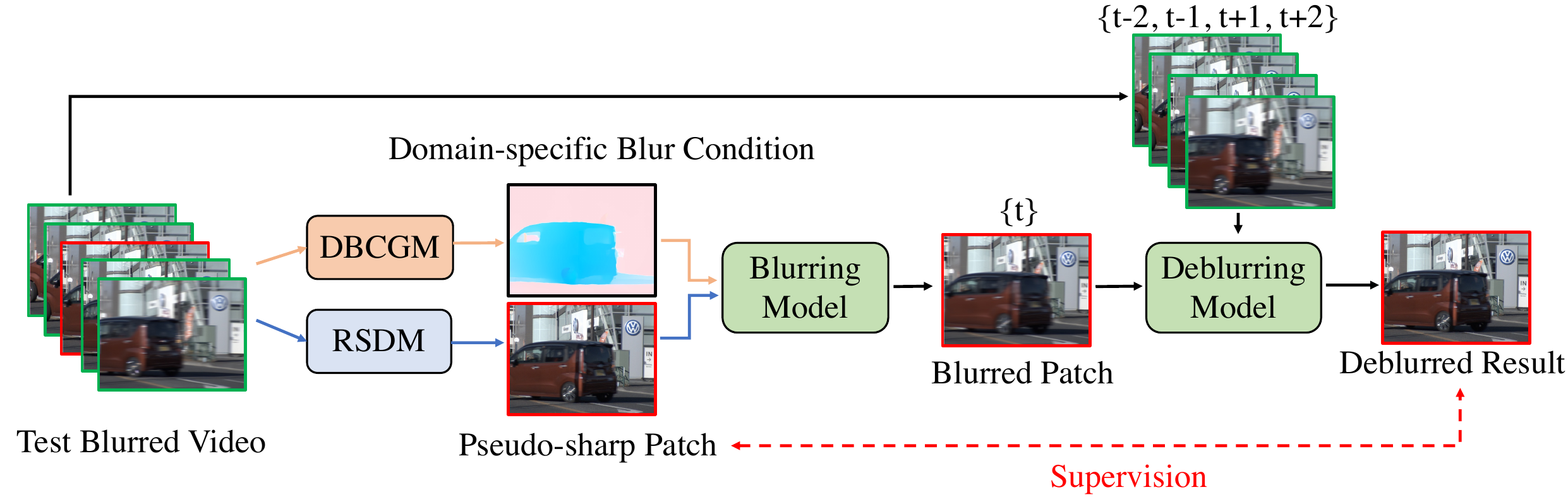}
\end{center}
\caption{Pipeline of the proposed domain adaptation scheme. Given a blurred video in the test domain, we use the Relative Sharpness Detection Module (RSDM) to extract relatively sharp patches to be pseudo-sharp patches and the Domain-adaptive Blur Condition Generation Module (DBCGM) to generate domain-specific blur conditions for blurring the pseudo-sharp patches using ID-Blau~\cite{ID-Blau}. Finally, the pseudo-sharp and blurred pairs are used to update a deblurring model for domain adaptation.     
}
\label{fig:pipeline}
\end{figure}

\section{Proposed Method}
We proposed a domain adaptation scheme for video deblurring based on ID-Blau~\cite{ID-Blau}, a diffusion-based blurring model that utilizes controllable blur conditions to generate corresponding blurred images. As shown in Figure~\ref{fig:pipeline}, considering blurred and sharp pairs are unavailable during inference, we generate domain-adaptive training pairs from blurred videos to calibrate a deblurring model for the target domain.
First, we propose a Relative Sharpness Detection Module (RSDM) to extract relatively sharp patches from blurred videos. These patches are treated as pseudo-sharp images, which are then blurred using ID-Blau to create pseudo-training pairs for updating deblurring models. 
In ID-Blau, a blur pattern is represented using pixel-wise blur orientations and magnitudes as a blur condition in a continuous blur condition field. Simply using ID-Blau with randomly sampled blurred conditions to produce blurred images may not help a deblurring model in unseen domains. 
Therefore, we proposed a Domain-adaptive Blur Condition Generation Module (DBCGM) to create domain-specific blur conditions for ID-Blau. It allows ID-Blau to generate blurred images adaptive to the target domain. Finally, the generated pseudo-sharp and blurred pairs are used to fine-tune a deblurring model for better performance. 

\subsection{Relative Sharpness Detection Module (RSDM)}
The proposed RSDM aims to search for relatively sharp patches in blurred videos. These patches are regarded as pseudo-sharp patches for domain adaptation. To achieve this, we propose a Blur Magnitude Estimator (BME) to predict a blur magnitude map for a blurred image, where the blurriness degree for each pixel is estimated. 

\begin{figure}[t!]
\begin{center}
\includegraphics[width=1\columnwidth]{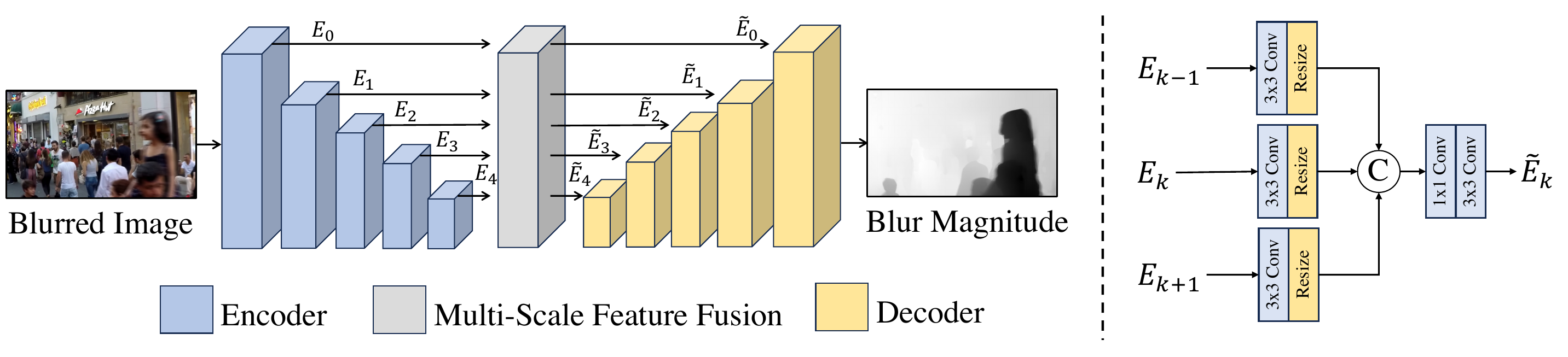}
\end{center}
\caption{The left figure shows the architecture of the Blur Magnitude Estimator (BME), which comprises a five-stage encoder-decoder design with Multi-Scale Feature Fusion (MSFF). The right figure is the architecture of MSFF. $E_{k}$ denotes the output of the encoder at stage $k$, and $\tilde{E}_{k}$ denotes its output after MSFF.
}
\label{fig:magnitude estimator}
\end{figure}

As shown in Figure~\ref{fig:magnitude estimator}, the BME is a five-stage encoder-decoder network combined with Multi-Scale Feature Fusion (MSFF) proposed in~\cite{MIMO}, as

\begin{equation}
\tilde{E}_k = 
\begin{cases} 
MSFF(E_{k-1}, E_{k}, E_{k+1}) & \text{if }  k = 1, 2, 3, \\
MSFF(E_{k}, E_{k+1}, E_{k+2}) & \text{if }  k = 0,  \\
MSFF(E_{k}, E_{k-1}, E_{k-2}) & \text{if }  k = 4,
\end{cases}
\end{equation}
where $E_{k}$ is the output of the $k$-th encoder layer and $\tilde{E}_k$ is its fusion result. MSFF includes a $3\times 3$ convolutional layer plus resizing all the input features to the size of $E_k$. After resizing, the multi-scale features are concatenated, followed by a $1\times 1$ and a $3\times 3$ convolutional layers.


%
To optimize the BME, we choose the GoPro~\cite{Nah_2017_CVPR} dataset, which synthesizes blurred images by accumulating consecutive sharp frames from a high-speed camera as 
\begin{equation}
    B = g(\frac{1}{T}\int_{t=1}^{T}H(t)dt) \simeq g(\frac{1}{N_s}\sum_{n=1}^{N_s}H\left[n\right]),
\label{eqa:accumulaiton}
\end{equation}
where $B$, $H$, $T$, $N_s$, and $g$ respectively denote the generated blurred image, the sharp images captured by a high-speed camera, the exposure time, the number of sampled sharp images, and the camera response function. Here, the center sharp image $H[\frac{N+1}{2}]$ is chosen to be the ground-truth sharp image $S$ corresponding to the generated blurred image $B$.

\begin{figure}[t!]
\begin{center}
\includegraphics[width=1\columnwidth]{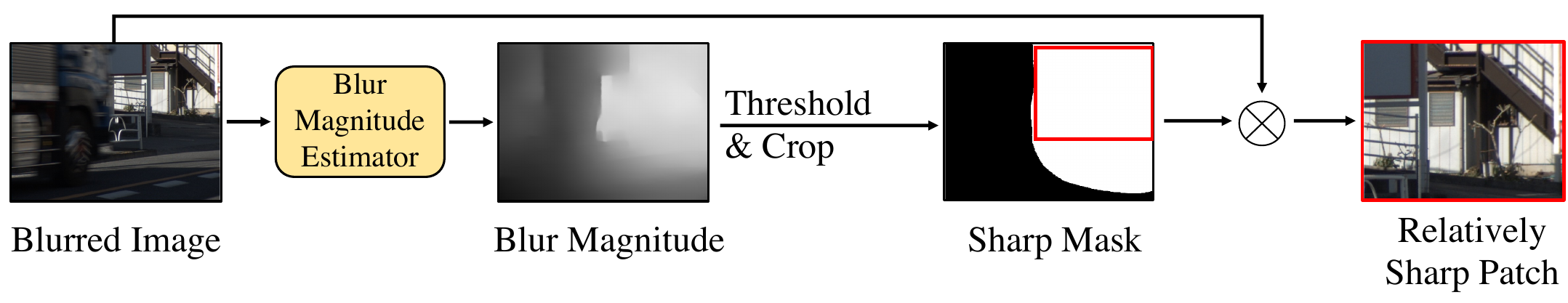}
\end{center}
\caption{Illustration of the Relative Sharpness Detection Module. We use the Blur Magnitude Estimator to obtain the blur magnitudes for a blurred image and crop a relatively sharp patch based on an adaptive sharpness threshold.}
\label{fig:rsdm}
\end{figure}

Considering continuous motion during the exposure process causes various degrees of blurriness, we characterize the continuous exposure as motion trajectories by accumulating optical flows from the sharp image sequence $\textbf{H}=\{H\left[1 \right], ..., H\left[N \right]\}$ and aggregating them as    
\begin{equation}
    \mathcal{F} = \sum_{n=1}^{N-1}\frac{f(H\left[n\right], H\left[n+1\right]) - f(H\left[n\right], H\left[n-1\right])}{2}, 
\end{equation}
where $f$ is a pre-trained optical flow network. Here, we adopt the method in~\cite{RAFT} for obtaining the optical flows. The calculated motion trajectory map has two components: $u$ and $v$, representing the average horizontal and vertical motion trajectories as $\mathcal{F}=[u; v] \in\mathbb{R}^{H \times W \times 2}$. 
Lastly, the blur magnitude map $G\in\mathbb{R}^{H \times W}$ corresponding to $B$ can be obtained by 
\begin{equation}
    G = \frac{1}{\tau}\sqrt{u^2 + v^2},
\label{eq:gt_magnitue}
\end{equation}
where $\tau$ serves as a normalization term, set to the maximum value of the blur magnitudes, and thus $G\in [0,1]$. To this end, a dataset $\{B_k, G_k\}^{K}_{k=1}$ is constructed to train the BME, where $K=2,103$ for the GoPro training set. 

Next, we take the trained BME to obtain the blur magnitudes for every frame in the testing blurred video to sift through relatively sharp patches. As shown in Figure~\ref{fig:rsdm}, given a blurred video $V^{(i)}$ in the test set, BME predicts a blur magnitude map $M^{(i)}_{t}\in\mathbb{R}^{H \times W}$ for each blurred frame $V^{(i)}_{t}$ as 
\begin{equation}
    M^{(i)}_{t} = BME(V^{(i)}_{t}),
    \label{eq:BME}
\end{equation}
where $t$ denotes the frame index.
To crop a relatively sharp patch from $V^{(i)}_{t}$, we use an adaptive sharpness threshold $\eta^{(i)}$ to binarize $M^{(i)}_{t}$ with magnitude larger than $\eta^{(i)}$ set to $1$ and obtain the sharp mask. We then crop a sharp patch by the regionprop library~\cite{regionprop} based on the mask to generate a relatively sharp patch $\tilde{S}^{(i)}_{t}$ of size $256\times256$. Here, $\eta^{(i)}$ is determined to ensure that the number of relatively sharp patches extracted from the video $V^{(i)}$ reaches $r\%$ of the number of total frames in $V^{(i)}$, meaning the top $r\%$ relatively sharp patches in the video. In our work, $r$ is set to $20$. 
%

%

%

\begin{figure}[t!]
\begin{center}
\includegraphics[width=1\columnwidth]{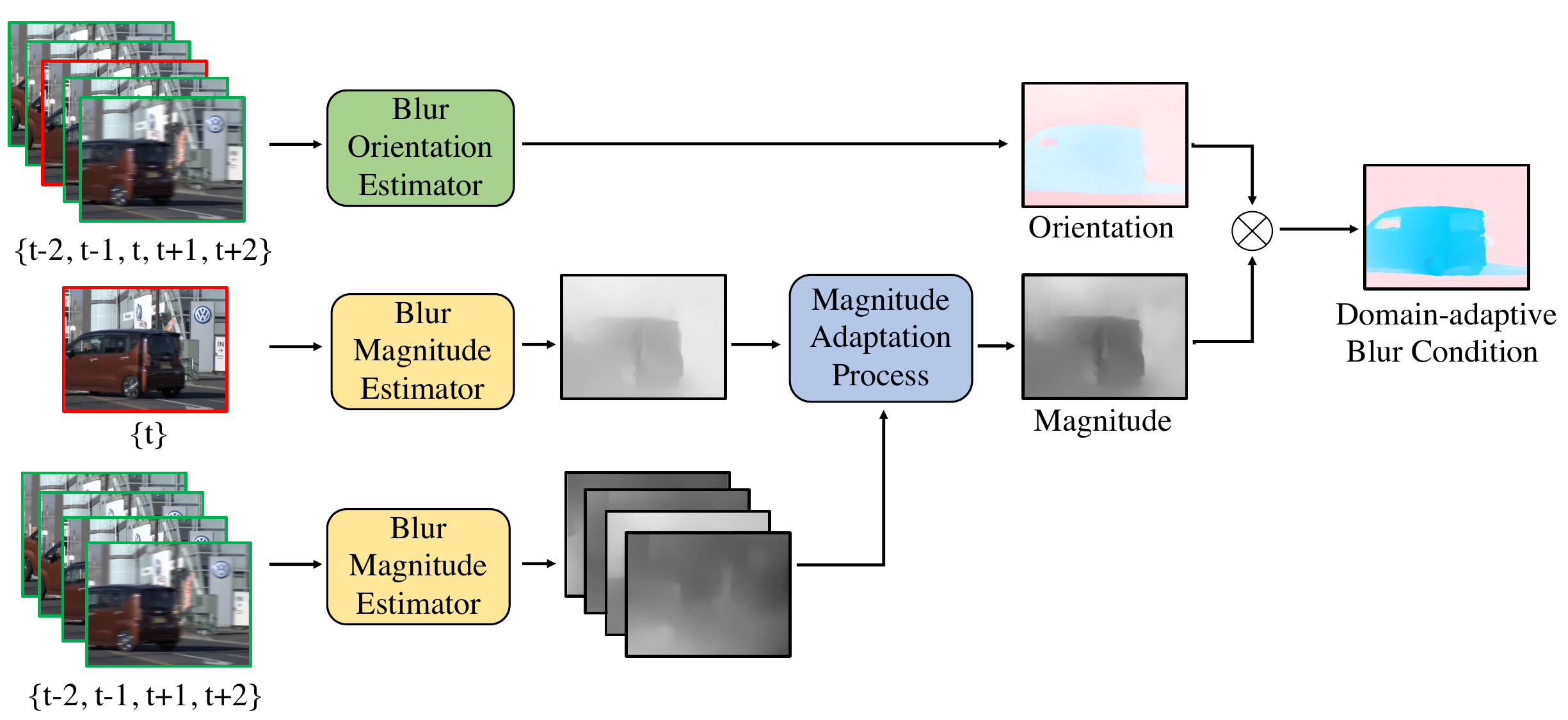}
\end{center}
\caption{Illustration of Domain-adaptive Blur Condition Generation Module. Given a pseudo-sharp patch and the collocated patches in its neighboring frames, we use the Blur Orientation Estimator to generate domain-specific blur orientations and the Blur Magnitude Estimator to generate domain-specific blur magnitudes. The blur magnitudes estimated from neighboring patches are used to modulate the pseudo-sharp patch by the Magnitude Adaptation Process. In the end, the domain-specific blur orientations and magnitudes are used for blurring.     
}
\label{fig:bcgm}
\end{figure}

\subsection{Domain-adaptive Blur Condition Generation Module (DBCGM)}
With the selected pseudo-sharp patches, we utilize a blurring model to generate their blurred versions for fine-tuning a deblurring model. In this work, we use ID-Blau~\cite{ID-Blau}, a conditional diffusion-based blurring model that takes a sharp image $S$ and a controllable blur condition map $C=(x,y,z)\in\mathbb{R}^{H \times W \times 3}$ to synthesize a blurred image $B$ as
\begin{equation}
    B = \textbf{ID-Blau}(S,C),
\end{equation}
where $x$, $y$, and $z\in\mathbb{R}^{H \times W}$ denote horizontal and vertical blur orientations, and blur magnitudes, respectively.
Although randomly generated blur conditions can be used to produce blurred images by ID-Blau, they do not conform to the test data distribution in the target domain, which means the generated blurred images do not agree with the blur patterns that exist in test videos, leading to little performance gain for deblurring models.
Therefore, as shown in Figure~\ref{fig:bcgm}, we propose a Domain-adaptive Blur Condition Generation Module (DBCGM) that can generate domain-specific blur conditions by leveraging motion cues implicitly provided in blurred videos during inference. 

In a blurred video, continuous motion across consecutive frames reveals motion blur trajectories during capturing, and the degree of blurriness implies the blur intensities during exposure. Therefore, the proposed DBCGM includes a Blur Orientation Estimator (BOE) and the previously mentioned BME to generate domain-specific blur conditions for blurring.
First, given a pseudo-sharp patch $\tilde{S}^{(i)}_{t}$ in the $t$-th frame of the $i$-th video, we take it and its collocated patches from the two previous and two next frames $\{\tilde{S}^{(i)}_{t-2}, ...,\tilde{S}^{(i)}_{t+2}\}$ to calculate the motion trajectory map ${\tilde{F}^{(i)}}_{t}=[\tilde{u}; \tilde{v}] \in\mathbb{R}^{H \times W \times 2}$ as
\begin{equation}
    {\mathcal{\tilde{F}}^{(i)}}_{t} = {\sum_{n=-2}^{1}f(\tilde{S}^{(i)}_{t+n}, \tilde{S}^{(i)}_{t+n+1})}, 
\end{equation}
where $f$ denotes the pre-trained optical flow estimator~\cite{RAFT}.
Next, we obtain the domain-specific blur orientations $\tilde{O}^{(i)}_{t}\in\mathbb{R}^{H \times W \times 2}$ tailored for $\tilde{S}^{(i)}_{t}$ as    
\begin{equation}
    \tilde{O}^{(i)}_{t} = \frac{\mathcal{\tilde{F}}^{(i)}_{t}}{\sqrt{\tilde{u}^2 + \tilde{v}^2}}.
\end{equation}
%
To generate domain-adaptive blur magnitudes for $\tilde{S}^{(i)}_{t}$, we first estimate blur magnitudes of $\tilde{S}^{(i)}_{t}$ by the BME as ${M}^{(i)}_{t} = BME(\tilde{S}^{(i)}_{t})$. Since $\tilde{S}^{(i)}_{t}$ is relatively sharp, we intend to modulate its blur magnitudes by considering blur patterns rendered in its neighboring collocated patches $\{\tilde{S}^{(i)}_{t-2}, \tilde{S}^{(i)}_{t-1}, \tilde{S}^{(i)}_{t+1}, \tilde{S}^{(i)}_{t+2}\}$. 
To achieve this, we use the average blur magnitudes of neighboring collocated patches to adjust ${M}^{(i)}_{t}$ by the Magnitude Adaptation Process as
\begin{equation}
    \tilde{M}^{(i)}_{t} = \text{Norm}(M^{(i)}_{t}) \cdot \text{Avg}(M^{(i)}_{t-2}, M^{(i)}_{t-1}, M^{(i)}_{t+1}, M^{(i)}_{t+2}),
    \label{eq:fusion}
\end{equation}
where $M^{(i)}_{t}$ is normalized in the range of $[0, 1]$ and multiplied by the blur magnitude average to generate domain-specific blur magnitudes $\tilde{M}^{(i)}_{t}\in\mathbb{R}^{H \times W}$.
At last, a domain-specific blur condition $\tilde{C}^{(i)}_{t}$ that combines blur orientations $\tilde{O}^{(i)}_{t}\in\mathbb{R}^{H \times W \times 2}$ and magnitudes $\tilde{M}^{(i)}_{t}\in\mathbb{R}^{H \times W}$ 
is used to blur $\tilde{S}^{(i)}_{t}$ by
\begin{equation}
    \tilde{B}^{(i)}_{t} = \textbf{ID-Blau}(\tilde{S}^{(i)}_{t},\tilde{C}^{(i)}_{t}),
\end{equation}
where $\tilde{B}^{(i)}_{t}$ is the generated blurred patch. The pseudo-training pair $\{\tilde{B}^{(i)}_{t}, \tilde{S}^{(i)}_{t}\}$ is then utilized to update a deblurring model for domain adaptation. Ultimately, we collect these pairs from the total $N$ blurred videos in the target domain to fine-tune a deblurring model for domain adaptation.

\subsubsection{Loss Function:}
In the proposed scheme, we only need to optimize the BME. We use the L1 loss,
\begin{equation}
\mathcal{L} = \mathcal{L}_\mathrm{1}(M, G),
\end{equation}
where $M$ is the predicted blur magnitudes, and $G$ is the calculated blur magnitudes using Equation~\ref{eq:gt_magnitue}, used as the ground truth.

\begingroup
\renewcommand\thefootnote{}\footnotetext{The authors from the universities in Taiwan completed the experiments on the datasets.}
\addtocounter{footnote}{-1}
\endgroup
\section{Experiments}
\subsection{Implementation Details}

\subsubsection{Blurring Model}
We choose a diffusion-based blurring ID-Blau~\cite{ID-Blau} as the blurring network to generate blurred images. 
Note that we use ID-Blau with its original training settings. 
For optimizing the BME, we use the Adam optimizer~\cite{kingma2017adam} with the initial learning rate of $1e^{-3}$,  gradually decayed to $1e^{-4}$ by the cosine annealing strategy. We resize images to $320\times320$ and adopt random flipping and rotation for data augmentation with a batch size of $16$ for training the model $50$ epochs.        
We use the GoPro training set~\cite{Nah_2017_CVPR}, which contains $22$ training videos with $2,103$ blurred and sharp image pairs for optimizing ID-Blau and BME.

\subsubsection{Video Deblurring Models}
We adopt four state-of-the-art video deblurring models, including ESTRNN~\cite{zhong2022real}, MMP-RNN~\cite{wang2022MMP}, DSTNet~\cite{Pan_2023_CVPR}, and Shift-Net~\cite{Li_2023_CVPR} to validate the effectiveness of the proposed domain adaptation method.
We regard the GoPro training set as the source domain and five real-world deblurring datasets as target domains, including BSD-1ms8ms~\cite{zhong2022real}, BSD-2ms16ms~\cite{zhong2022real}, BSD-3ms24ms~\cite{zhong2022real}, RealBlur~\cite{rim_2020_ECCV}, and RBVD~\cite{chao2022} test sets.
The BSD test set has three subsets: 1ms8ms, 2ms16ms, and 3ms24ms, indicating two exposure times: the former of which is for sharp images and the latter of which is for blurred ones. Each subset contains $20$ videos with $3,000$ blurred images. 
The RealBlur test set consists of $50$ videos with $980$ blurred images, taken in low-light environments.
The RBVD test dataset consists of $7$ videos with $246$ blurred images, encompassing outdoor scenes, indoor scenes, and high-frequency charts.
In the domain adaptation process, we fine-tune each deblurring model for $10$ epochs on pseudo-training pairs generated by the proposed RSDM and DBCGM, as we use each model's original loss functions for training.

\begin{table*}[t!]
\centering
\setlength{\tabcolsep}{0.5mm}
\caption{Evaluation results on the deblurring datasets, including BSD~\cite{zhong2022real}, RealBlur~\cite{rim_2020_ECCV}, and RVBD~\cite{chao2022}, where ``Baseline'' and ``+Ours'' denote the video deblurring performances w/o or w/ our domain adaptation method.}
\scriptsize
\begin{tabular}{cc |lc| lc| lc| lc| cc}
\hline\hline
& & \multicolumn{2}{c|}{\bf{BSD-1ms8ms}} & \multicolumn{2}{c|}{\bf{BSD-2ms16ms}} & \multicolumn{2}{c|}{\bf{BSD-3ms24ms}} & \multicolumn{2}{c|}{\bf{RealBlur}} & \multicolumn{2}{c}{\bf{RBVD}} \\
\multicolumn{2}{c|}{Model}  & PSNR   & SSIM   & PSNR  & SSIM & PSNR  & SSIM & PSNR  & SSIM & PSNR  & SSIM  \\ \hline
\multirow{2}{*}{ESTRNN} & Baseline &   25.57    &  0.747   &  24.64   &   0.726 &   26.01    &  0.748   & 25.87  & 0.773  & 24.47  & 0.725 \\ & +Ours &   \bf 29.44    &  \bf 0.843   &  \bf28.36  &   \bf0.820    &  \bf28.32  &   \bf0.810 &  \bf27.64  &   \bf0.816 & \bf26.83 & \bf 0.764 \\ \hline

\multirow{2}{*}{MMP-RNN} & Baseline &  21.63    &  0.620   &  21.26   &   0.605 &   22.74    &  0.597   &   24.65  & 0.639 &   22.81  & 0.780 \\ & +Ours &   \bf29.17   &  \bf0.797   &  \bf26.95  &   \bf0.750 &   \bf26.77   &  \bf0.707   &  \bf 27.69 &   \bf0.663 & \bf25.81 & \bf0.822 \\ \hline

\multirow{2}{*}{DSTNet} & Baseline &   25.42    &  0.821   &  23.50   &   0.760 &   24.68    &  0.788   &  26.57   &   0.750 &  23.15   &   0.768 \\ & +Ours &   \bf28.69   &  \bf0.868   &  \bf27.11  &   \bf0.830 &   \bf26.69   &  \bf0.838   &  \bf 27.74  &   \bf0.798 & \bf25.66 & \bf0.808 \\ \hline

\multirow{2}{*}{Shift-Net} & Baseline &   25.00    &  0.837   &  23.75   &   0.807 &   24.98    &  0.819   &  26.01  &   0.797 &  23.98  &   0.870 \\ & +Ours &   \bf 28.75   &  \bf0.888   &  \bf26.31  &   \bf0.854 &  \bf26.92   &  \bf0.852   &  \bf27.71  &   \bf0.854 & \bf25.35 & \bf0.891 \\ \hline

\hline\hline
\end{tabular}
\label{Tab:Performance}
\end{table*}

\subsection{Experimental Results}

\subsubsection{Quantitative Analysis}
We compare the performance of four video deblurring models, including ESTRNN~\cite{zhong2022real}, MMP-RNN~\cite{wang2022MMP}, DSTNet~\cite{Pan_2023_CVPR}, and Shift-Net~\cite{Li_2023_CVPR}, on five real-world deblurring datasets with (+Ours) or without (Baseline) using the proposed domain adaptation scheme in Table~\ref{Tab:Performance}. It shows that our method can consistently and significantly improve the performance for these video deblurring models by $4.61$dB, $3.90$dB, $2.57$dB, $1.92$dB, and $2.31$dB in PSNR on average on the BSD-1ms8ms, BSD-2ms16ms, BSD-3ms24ms, RealBlur and RBVD test sets, respectively. Particularly, our method can obtain up to $3.87$dB, $7.54$dB, $3.61$dB, and $3.75$dB performance gains for deblurring models: ESTRNN, MMP-RNN, DSTNet, and Shift-Net, respectively.
These experimental results show that our method can generate domain-adaptive training pairs to effectively fine-tune deblurring models originally trained with a synthetic dataset for those real-world blurred videos.     

\subsubsection{Qualitative Analysis}
\begin{figure}[t!]
\begin{center}
\includegraphics[width=1\columnwidth]{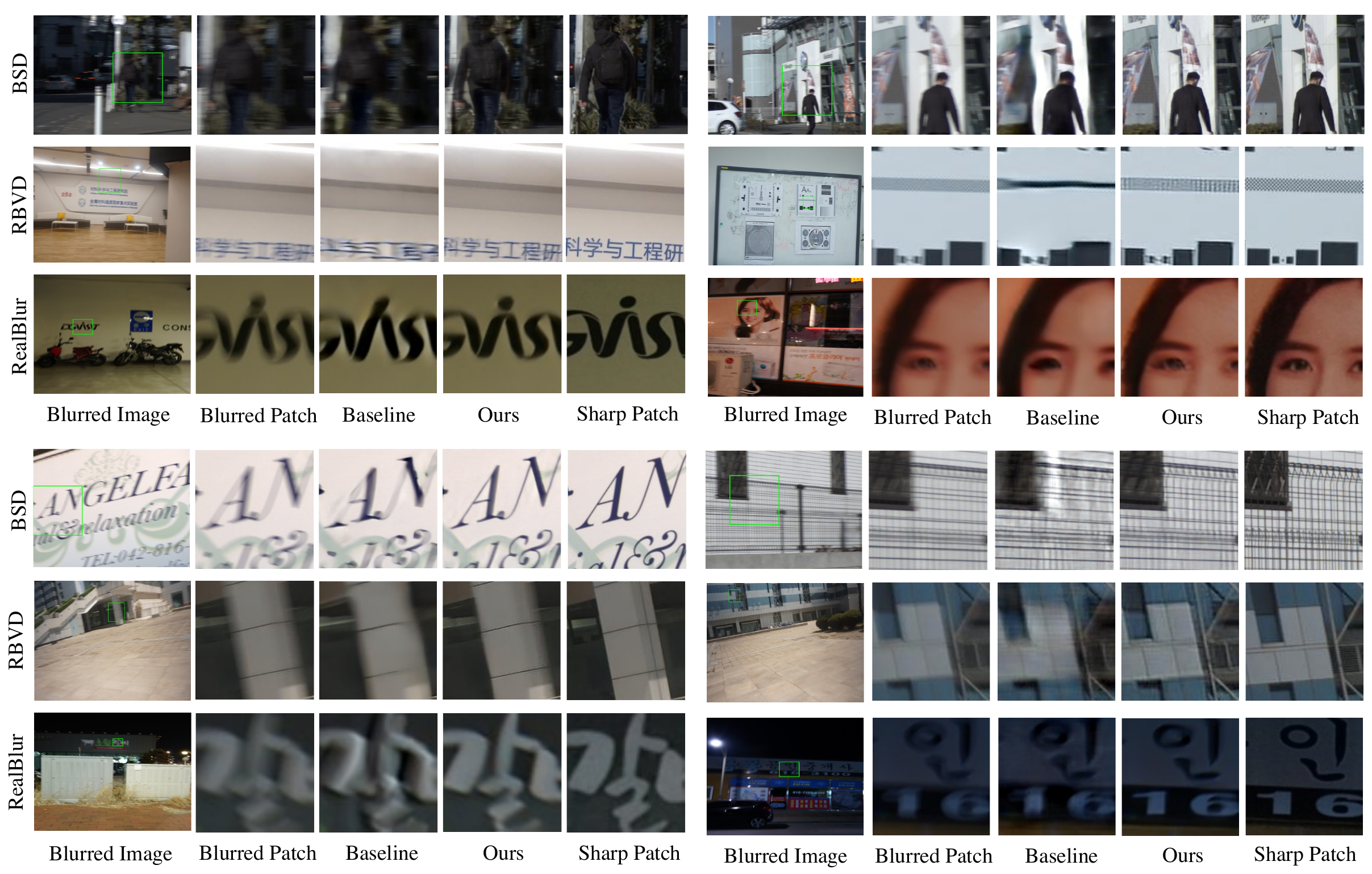}
\end{center}
\caption{Qualitative results on the BSD datasets, RBVD dataset and RealBlur dataset with DSTNet(top left), MMP-RNN(top right), Shift-Net(bottom left), and ESTRNN(bottom right).
}
\label{fig:deblur_result}
\end{figure}

In Figure~\ref{fig:deblur_result}, we compare deblurred results with or without using our method on several blurred video frames chosen from the BSD test set. Our domain-adaptive scheme can significantly improve the visual quality of the deblurred results, exemplified by the regions of the face, text, and pedestrians.
Figure~\ref{fig:deblur_result} shows comparisons of the deblurred results with or without using our method on blurred examples on the RBVD and RealBlur test sets. The visual quality of deblurred results using our method is significantly better, exemplified by the regions of the pillar, text, and building.
These visualization results show that our method can truly help boost the deblurring performance compared to the baseline methods. It also attests to the effectiveness of the proposed domain adaptation scheme to produce domain-adaptive training pairs for model fine-tuning in the target domain.



\subsection{Ablation studies}
To analyze the impact of the proposed components in our domain adaptation method, we adopt ESTRNN~\cite{zhong2022real} as the tested deblurring model and analyze its deblurring performances on the BSD-1ms8ms dataset with various ablative settings. 
First, we discuss the effectiveness of the RSDM and DBCGM. Next, we discuss the designs in RSDM and DBCGM. Lastly, we compare the proposed method with the existing domain adaptation method~\cite{Liu_2022_BMVC} for video deblurring.   

\begin{table}[t!]
\centering
\caption{Ablations on the effectiveness of the RSDM and DBCGM. }
\scriptsize
\begin{tabular}{ccc|ccc|cc}
\hline\hline
& \multicolumn{2}{c|}{\bf{Pseudo-Sharp Patches}} & \multicolumn{3}{c|}{\bf{Blur Condition Generation}} & \bf{PSNR} & \bf{GAIN} \\
& Random   & RSDM(Ours)  & Random  & Optical-Flow & DBCGM(Ours) &   \\ \hline
 (a) & & & & & & 25.57 & +0.00 \\
 (b) & \Checkmark & & \Checkmark & & & 23.88 & -1.69 \\
 (c) &\Checkmark & & &  \Checkmark & &  25.51 & -0.06 \\
 (d) &\Checkmark & & &  &  \Checkmark &  29.01 & +3.44   \\ 
 (e) &&\Checkmark & \Checkmark &  &  & 24.32 & -1.25   \\ 
 (f) &&\Checkmark & & \Checkmark &  & 26.19 & +0.62   \\ 
 (g) &&\Checkmark & &  & \Checkmark &  \bf29.44 & \bf+3.87   \\ 
\hline\hline
\end{tabular}
\label{Tab:Ablation_Module}
\end{table}

\subsubsection{Effects of RSDM and DBCGM}
\begin{figure}[t!]
\begin{center}
\includegraphics[width=1\columnwidth]{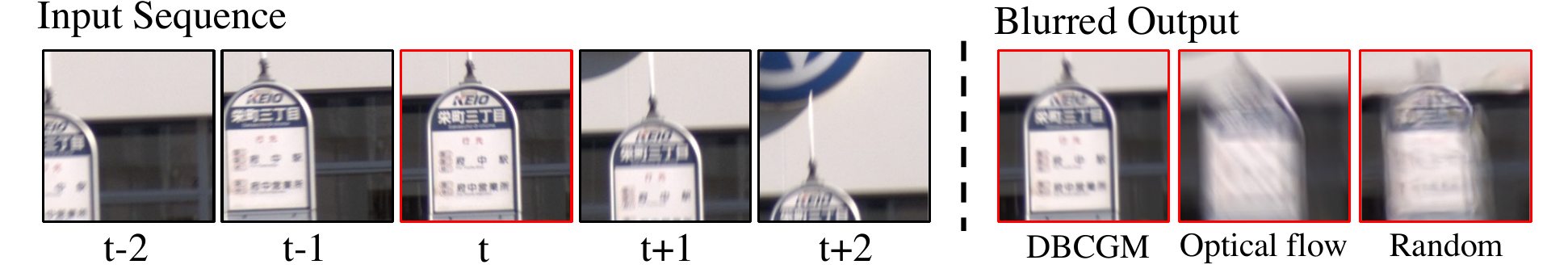}
\end{center}
\caption{Visualization of blurred images obtained using ID-Blau with different blur condition generation methods: DBCGM, optical flows, and random sampling.
}
\label{fig:reblur_vis}
\end{figure}

Table~\ref{Tab:Ablation_Module} compares the effectiveness of the RSDM and DBCGM to the video deblurring performance. Table~\ref{Tab:Ablation_Module}(a) shows the deblurred results on BSD-1m8ms using the GoPro pre-trained model, considered the baseline setting. 
First, we analyze the effectiveness of the proposed DBCGM with randomly selected patches in blurred videos to be pseudo-sharp patches (Table~\ref{Tab:Ablation_Module}(b), (c), and (d)). To blur pseudo-sharp patches using ID-Blau, we can use randomly sampled blur conditions (Table~\ref{Tab:Ablation_Module}(b)) or take optical flows as blur conditions (Table~\ref{Tab:Ablation_Module}(c)). As seen, these two cases do not improve the performance at all. In contrast, using the proposed DBCGM (Table~\ref{Tab:Ablation_Module}(d)) can largely improve the baseline by $3.44$dB, which demonstrates that even with random patches cropped from blurred video frames, the deblurring model can still exploit domain-specific blur conditions generated by the DBCGM to achieve better performance.    
Next, we analyze the effectiveness of the proposed RSDM. In Table~\ref{Tab:Ablation_Module}, we can compare the settings (b) vs. (e), (c) vs. (f), and (d) vs. (g), which shows that using the RSDM with different blur condition generation methods can all boost the deblurring model. Most of all, the combination of the RSDM and DBCGM achieves the best performance, achieving a $3.87$dB gain compared to the baseline.
%
In addition, we visualize the blurred results in Figure~\ref{fig:reblur_vis} using different blur condition generation methods. It shows that the proposed DBCGM can generate domain-specific blurred images more realistic and consistent with video frames of a moving scene than optical flows or randomly sampled blur conditions.

\begin{figure}[t!]
\begin{center}
\includegraphics[width=1\columnwidth]{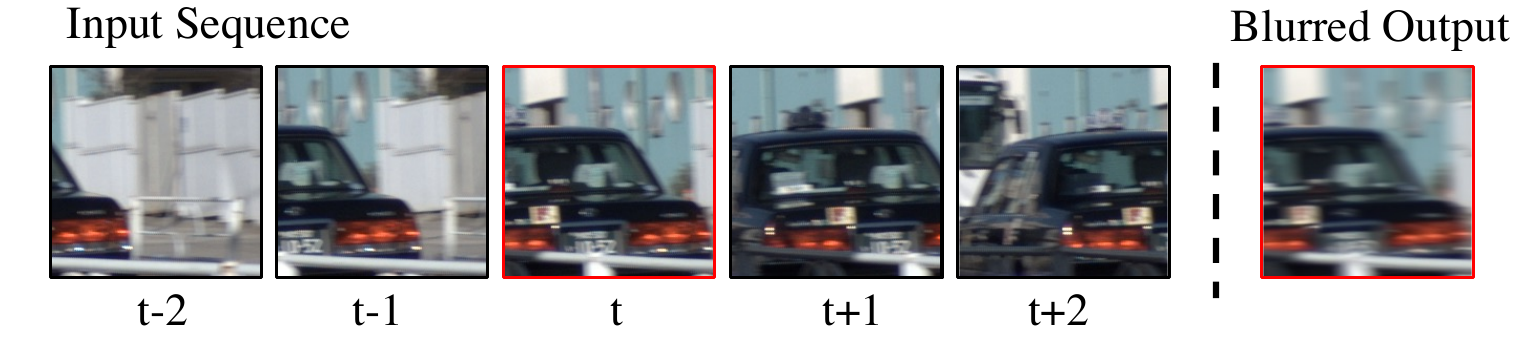}
\end{center}
\caption{Example of a blurred image generated from a blurry image using our method.
}
\label{fig:more_blur}
\end{figure}

\subsubsection{Effects of the adaptive sharpness threshold for the RSDM}
\begin{figure}[tb]
  \centering
  \begin{subfigure}{0.45\linewidth}
    \includegraphics[width=1\columnwidth]{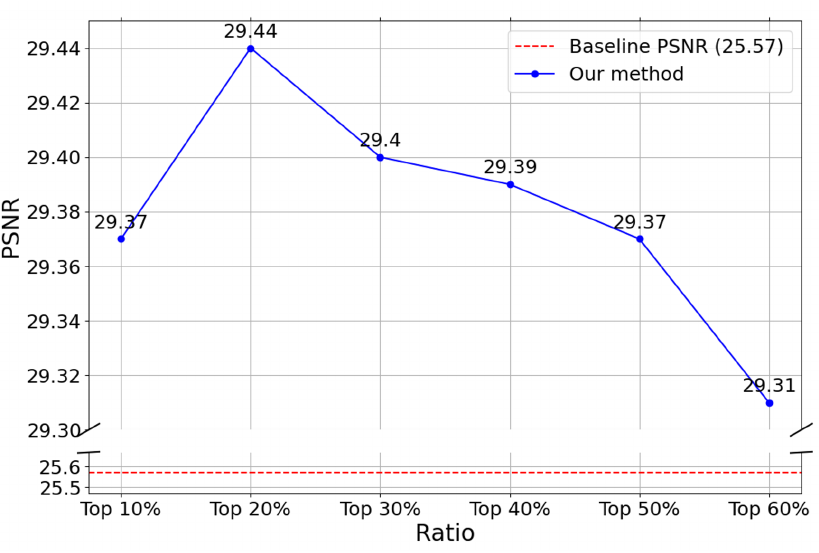}
    \caption{Performance of using different ratios for the threshold in RSDM}
    \label{fig:short-a}
  \end{subfigure}
  \hfill
  \begin{subfigure}{0.45\linewidth}
    \includegraphics[width=1\columnwidth]{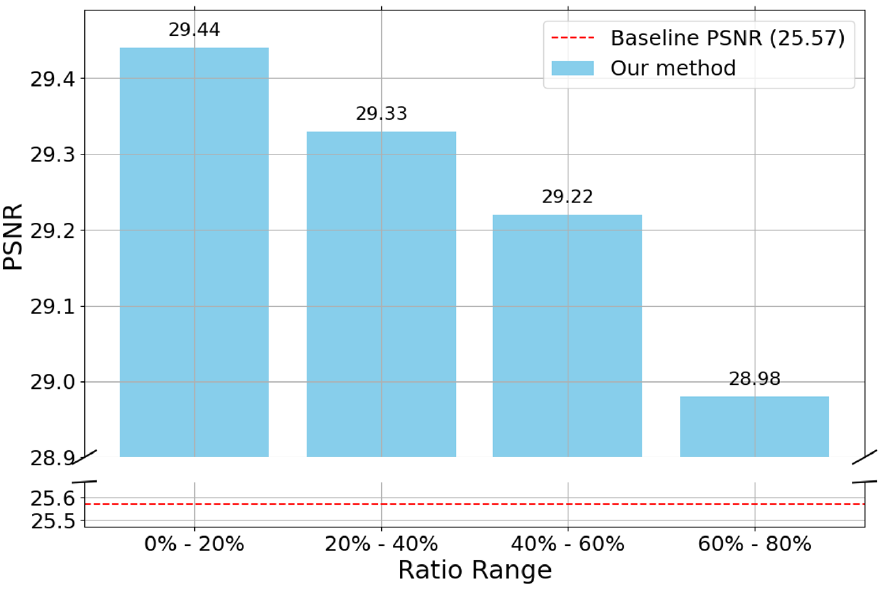}
    \caption{Performance of using different ratio ranges for picking pseudo-sharp patches}
    \label{fig:short-b}
  \end{subfigure}
  \caption{Illustration of the PSNR results when we set different threshold for RSDM.}
  \label{fig:threshold_ablation}
\end{figure}

The adaptive sharpness threshold $\eta^{(i)}$ is determined for the video $V^{(i)}$ in the RSDM to pick the top $r\%$ relatively sharp patches.
Choosing a smaller ratio can result in sharper patches to be pseudo-sharp patches. However, it may not provide sufficient training data for adaptation. In contrast, selecting a larger ratio would include more blurred patches as pseudo-sharp patches, leading to inferior pseudo-training pairs generated, where those blurred patches are too similar to their blurred versions.   
Therefore, we experiment to analyze the effect of using different ratios $r\%$ on the deblurring performance for adaptation.     
Figure~\ref{fig:short-a} shows the deblurring performance of using different ratios from $10\%$ ($3,000\times10\%=300$ patches) to $60\%$ ($3,000\times60\%=1,800$ patches). As seen, setting $r=20$ achieves the best performance. Choosing a larger ratio would include more inferior training pairs, providing less help for deblurring.
In addition, we conduct another experiment by using patches in different ratio ranges for blurring, including $0\%-20\%$, $20\%-40\%$, $40\%-60\%$, and $60\%-80\%$. For example, $20\%-40\%$ means we select the top $40\%$ relatively sharp patches but exclude the top $20\%$ ones. Based on Figure~\ref{fig:short-b}, using patches with more blur to generate pseudo-training pairs is not as beneficial as those with less blur to obtain quality pairs. That said, compared to the baseline, intentionally selecting patches with more blur in the proposed domain adaptation scheme still improves the deblurring performance. Figure~\ref{fig:more_blur} shows a blurred example using a patch with much blur.     



\subsection{Comparison with an existing domain adaptation method}
Considering little work regarding domain adaptation for video deblurring, we only compare the proposed method with~\cite{Liu_2022_BMVC}, which uses meta-learning to achieve domain adaptation. Table~\ref{Tab:Ablation_method} shows that our method significantly outperforms~\cite{Liu_2022_BMVC} on all the benchmark datasets.

\begin{table}[t!]
\centering
\setlength{\tabcolsep}{1mm}
\caption{Comparison between our method and Liu~\etal~\cite{Liu_2022_BMVC}.}
\begin{tabular}{c|ccccc}
\hline
Model & BSD-1ms8ms & BSD-2ms16ms & BSD-3ms24ms & RealBlur & RBVD \\ \hline
Baseline & 25.57 & 24.64 & 26.01 & 25.87 & 24.47 \\ 
Liu~\etal~\cite{Liu_2022_BMVC} & 25.58 & 24.53 & 25.15 & 26.12 & 24.83 \\ 
Ours & \bf29.44 & \bf28.36 & \bf28.32 & \bf27.64 & \bf26.83 \\
\hline
\end{tabular}
\label{Tab:Ablation_method}
\end{table}

\section{Conclusion}
We proposed a domain adaptation scheme for video deblurring based on a conditional diffusion-based blurring model to achieve test-time fine-tuning in unseen domains. To generate domain-adaptive training pairs for updating deblurring models during inference, we proposed two modules: the Relative Sharpness Detection Module (RSDM) and the Domain-adaptive Blur Condition Generation Module (DBCGM). The RSDM extracts relatively sharp patches from blurred input frames as pseudo-sharp images. The DBCGM generates domain-specific blur conditions for blurring the pseudo-sharp images. Lastly, the generated pseudo-sharp and blurred pairs are used to calibrate a deblurring model for better performance. 
Extensive experimental results have demonstrated that our approach can significantly improve state-of-the-art video deblurring methods, offering performance gains of up to 7.54dB on various real-world video deblurring datasets.

\section*{Acknowledgments}
This work was supported in part by the National Science
and Technology Council (NSTC) under grants 112-2221-EA49-090-MY3, 111-2628-E-A49-025-MY3, 112-2634-F002-005, 112-2221-E-004-005, 113-2923-E-A49-003-MY2, 113-2221-E-004-001-MY3 and 113-2622-E-004-001 This work was funded in part by Qualcomm through a Taiwan University Research Collaboration Project.
%
%
\bibliographystyle{splncs04}
\bibliography{main}
\end{document}